\documentclass[10pt,twocolumn,letterpaper]{article}

\usepackage[pagenumbers]{cvpr} 

\usepackage{array}
\usepackage{xcolor}
\usepackage{listings}
\usepackage{makecell}
\usepackage{subcaption}
\usepackage[dvipsnames]{xcolor}
\usepackage{pifont}
\usepackage{makecell}
\usepackage{multirow}
\definecolor{mytitlecolor}{rgb}{.729, .047, .184}
\usepackage{wrapfig}
\usepackage{subcaption}
\usepackage{amsmath}
\usepackage[accsupp]{axessibility}

\def\datasetName{CO\textcolor{mytitlecolor}{in}CO}

\definecolor{cvprblue}{rgb}{0.21,0.49,0.74}
\usepackage[pagebackref,breaklinks,colorlinks,allcolors=cvprblue]{hyperref}

\def\paperID{32450} 
\def\confName{CVPR}
\def\confYear{2026}

\title{Common \textcolor{mytitlecolor}{Inpainted} Objects \textcolor{mytitlecolor}{In-N-Out} of Context}
\author{
Tianze Yang\thanks{Equal contribution.} \quad
Tyson Jordan$^{*}$ \quad
Ruitong Sun$^{*}$ \quad
Ninghao Liu \quad
Jin Sun\\
University of Georgia\\
{\tt\small 
\{Tianze.Yang, tysonjordan, Ruitong.Sun, ninghao.liu, jinsun\}@uga.edu}
}

\begin{document}

\maketitle

\begin{abstract}
  We present Common Inpainted Objects In-N-Out of Context (\datasetName), a novel dataset addressing the scarcity of out-of-context examples in existing vision datasets. By systematically replacing objects in COCO images through diffusion-based inpainting, we create 97,722 unique images featuring both contextually coherent and inconsistent scenes, enabling effective context learning. Each inpainted object is meticulously verified and categorized as in- or out-of-context through Large Vision Language Model assessments. We demonstrate three key tasks enabled by \datasetName{}: (1) a fine-grained context reasoning approach that classifies objects as in- or out-of-context based on three criteria; (2) a novel Objects-from-Context prediction task that determines which new objects naturally belong in given scenes at both instance and clique level semantics, and (3) context-enhanced fake detection on state-of-the-art methods without fine-tuning. \datasetName{} provides a controlled testbed with contextual variations, establishing a foundation for advancing context-aware visual understanding in computer vision, including image forensics. Code and dataset are available at \url{https://co-in-co.github.io/}.
\end{abstract}


\section{Introduction}
\label{sec:intro}
\begin{figure}[t]
    \centering
    \includegraphics[width=0.45\linewidth]{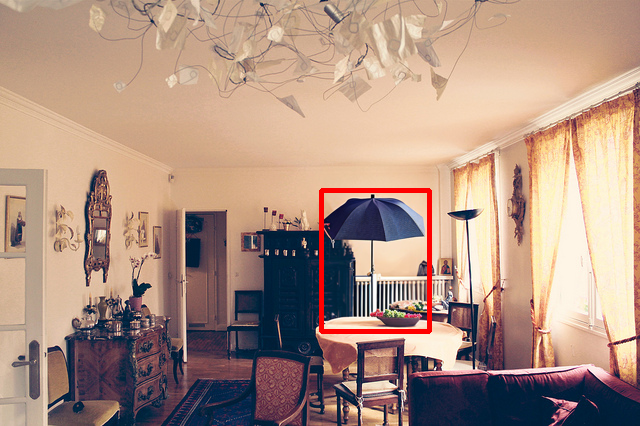}
    \includegraphics[width=0.45\linewidth]{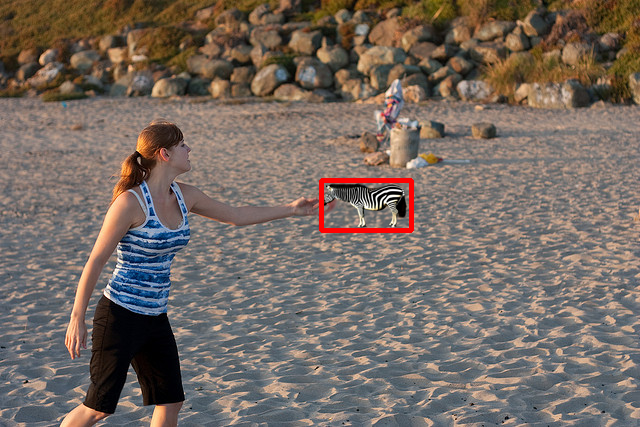}
    
    \vspace{2mm}
    
    \includegraphics[width=0.45\linewidth,height=1.2in]{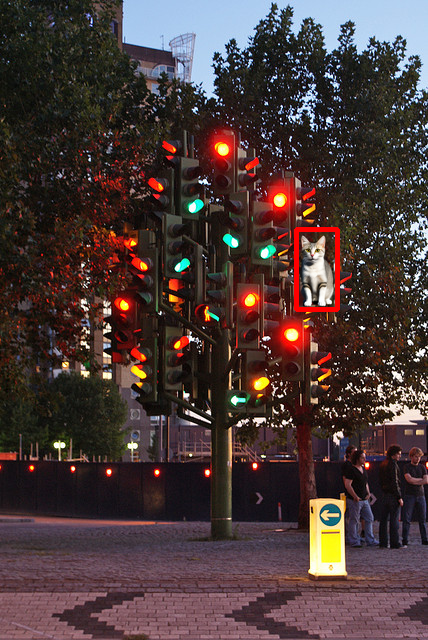}
    \includegraphics[width=0.45\linewidth,height=1.2in]{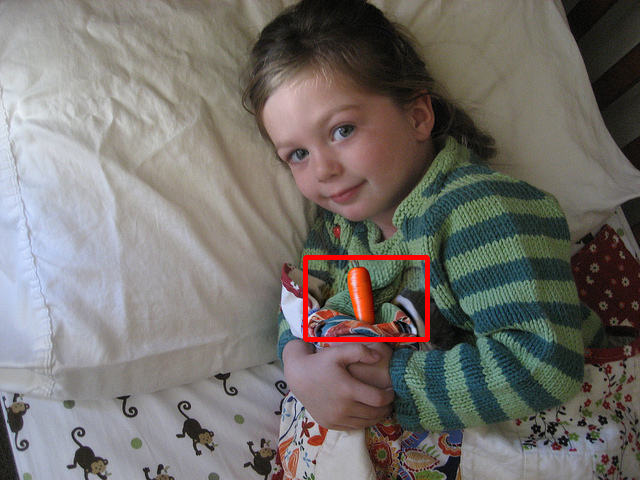}
    
    \vspace{2mm}
    
    \includegraphics[width=0.45\linewidth]{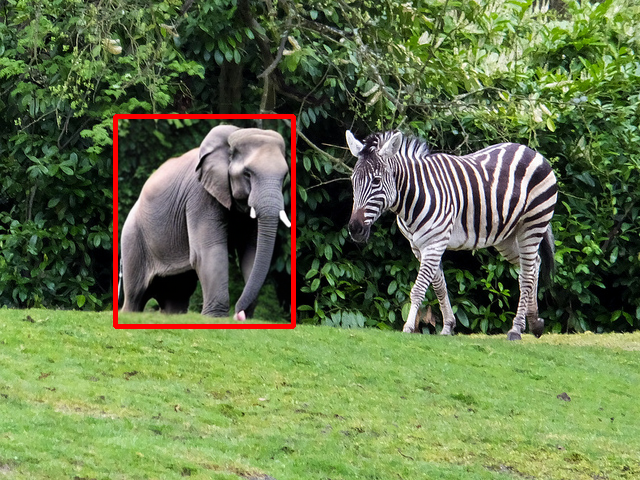}
    \includegraphics[width=0.45\linewidth]{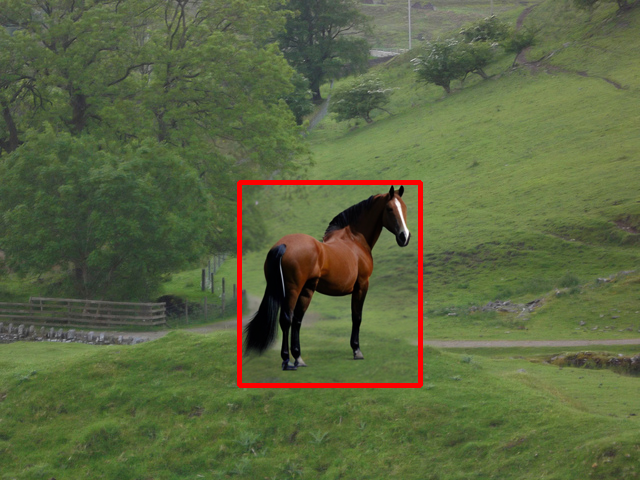}
    
    \caption{\datasetName{} contains a rich set of inpainted out-of-context objects (top two rows) and in-context objects (bottom row).}
    \label{fig:teaser}
\end{figure}

Context is fundamental to visual understanding~\cite{bar2004visual,oliva2007role,rabinovich2007objects,zhang2020putting,choi2012context, xiao2020noise}. When humans view a scene, we instinctively assess the coherence between objects and their environment. This context-based reasoning is essential for judging whether objects naturally belong in a scene~\cite{biederman1982scene}. Take a look at Figure~\ref{fig:teaser}—do the red-boxed objects seem reasonable in their contexts? Context reasoning can reveal out-of-place objects, providing a complementary cue for detecting manipulated content. By using context, humans assess whether objects appear in plausible settings—a horse in a grassy field with trees in the background appears perfectly natural, while a tiny zebra on a beach next to a person immediately raises suspicion due to its unusual size and location.

Learning context from data, however, is difficult. A significant challenge is that unusual scenes with out-of-context objects are rare in real life. Common computer vision datasets~\cite{lin2014microsoft, caesar2018coco,gupta2019lvis,kazemzadeh2014referitgame} primarily contain objects in their natural settings, resulting in a scarcity of examples with contextual violations. This presents a fundamental obstacle: how can we train machine learning models that require large data to recognize contextual inconsistencies when such examples are inherently uncommon?

To address this challenge, we propose a new dataset with both in-context and out-of-context objects through controlled image manipulation. By systematically replacing objects in existing real scenes, we generate the necessary examples of contextual violations while preserving the overall structure of the scene. We specifically build on top of the Common Objects in Context (COCO) dataset~\cite{lin2014microsoft} because it contains a diverse set of everyday scene photographs with rich image- and object-level annotations, offering an ideal starting point for contextual manipulation.

Using a diffusion-based inpainting model~\cite{rombach2022high}, 
we replace exactly one object per COCO image. This selective approach allows us to maintain the broader scene context while introducing precise, controlled variations in object-scene relationships. 
Our meticulous data creation pipeline and multi-step verification ensure high-quality inpainting results, and we apply a state-of-the-art Large Vision Language Model (LVLM) to classify each inpainted object as either contextually consistent or inconsistent with its scene.

\begin{figure*}[t]
    \centering
    \includegraphics[height=3.4in]{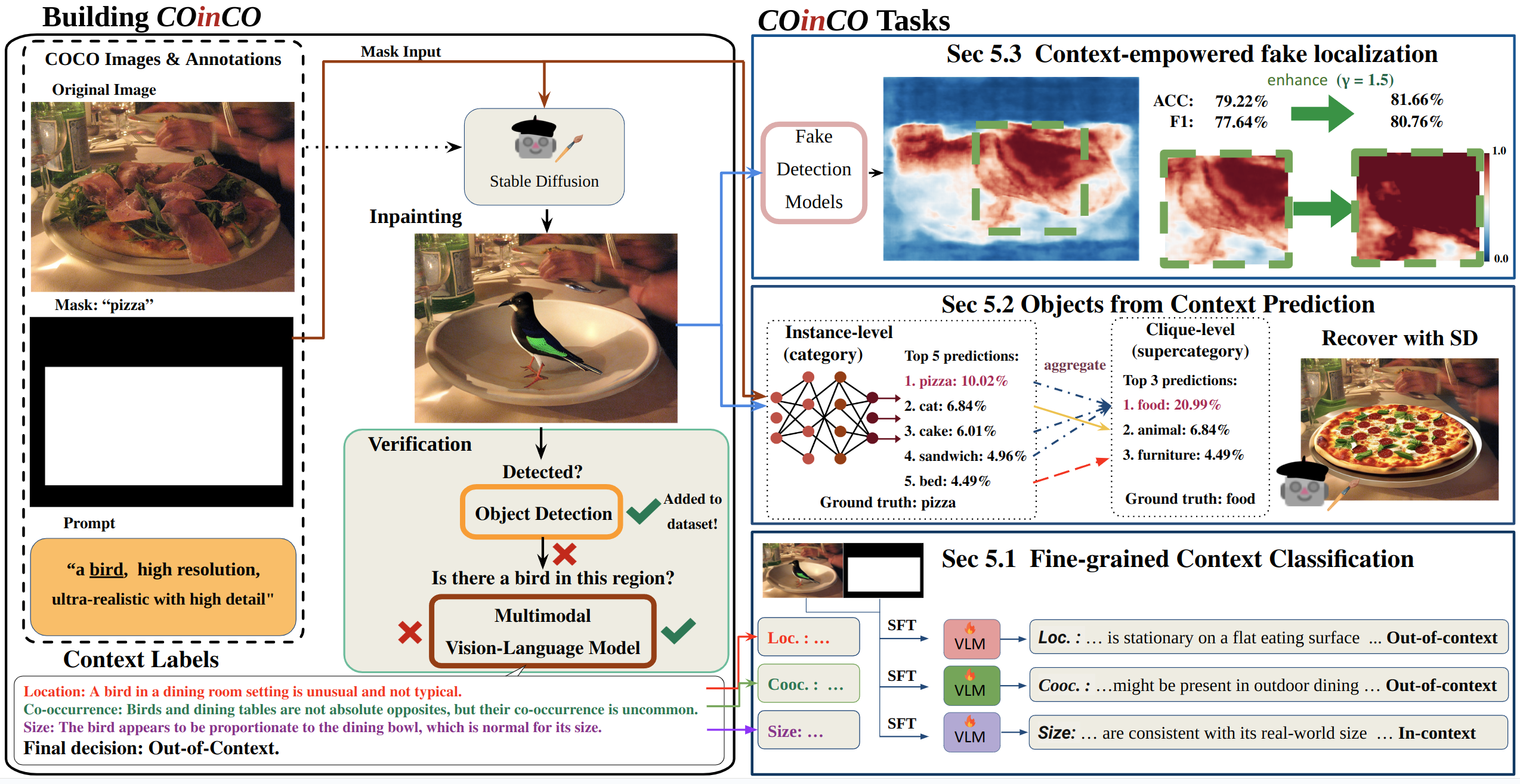}
    \caption{\datasetName{} pipeline. \textbf{Left: Building COinCO.} We start with COCO images and annotations. For each image, we randomly select an object and replace it using Stable Diffusion inpainting. We verify the inpainted object using object detection and a vision-language model. Successfully inpainted images are added to the dataset, while failed cases are regenerated and retested. Finally, each inpainted object is labeled as in-context or out-of-context based on three criteria: location, size, and co-occurrence. \textbf{Right: COinCO Tasks.} We demonstrate three downstream applications enabled by \datasetName. \textbf{Sec~5.1:} Fine-grained context classification evaluates objects on each criterion using vision-language models. \textbf{Sec~5.2:} Objects-from-context prediction identifies which object categories fit a given context at both instance and clique level semantics, and can recover the original object with an image generation model. \textbf{Sec~5.3:} Context-enhanced fake detection improves pretrained fake detectors by enhancing predictions in out-of-context regions.}
    \label{fig:pipeline}
    \vspace{-5pt}
\end{figure*}

We name our dataset \textit{Common \textcolor{mytitlecolor}{Inpainted} Objects \textcolor{mytitlecolor}{In-N-Out} of Context (\datasetName)}, emphasis on context for inpainted objects. It contains 97,722 inpainted images, each accompanied by rich annotations. 
Figure~\ref{fig:pipeline} shows the pipeline of our data creation and the downstream tasks.

\datasetName{} is the first dataset that features inpainted objects with diverse context annotations, enabling advancements in tasks including fine-grained context classification, object-from-context prediction, and fake detection (as our objects are inpainted by a generative model). 

\medskip{}
\noindent Our main contributions are:\\
\vspace{1pt}
\noindent 1. We introduce a novel, large-scale dataset of partially manipulated images that enhances COCO, featuring strategically inpainted objects that are either contextually coherent or inconsistent with their scenes.\\
\vspace{1pt}
\noindent 2. We introduce a fine-grained context reasoning task in \datasetName{} based on three criterion-specific dimensions and train efficient student models through distillation from a 72B LVLM teacher. These models provide modular, interpretable, and high-accuracy context reasoning for inpainted and real images.\\
\vspace{1pt}
\noindent 3. We introduce a novel task, object-from-context, which aims to predict instance- and clique-level semantic categories for possible objects that fit the given context.\\
\vspace{1pt}
\noindent 4. We show that context can be effectively integrated into fake detection pipelines, substantially improving state-of-the-art image forensics methods without fine-tuning.


\section{Related Work}
\label{sec:related}

\noindent\textbf{COCO and its extensions}.
The Common Objects in Context (COCO) dataset~\cite{lin2014microsoft} is a widely used benchmark for various computer vision tasks, including object detection, instance segmentation, and captioning. Over the years, several extensions have been proposed.
Some datasets primarily focus on \textit{expanding annotations} without modifying image content. COCO-WholeBody refines human keypoint detection by adding facial, hand, and foot keypoints~\cite{jin2020whole}, while LVIS introduces a long-tailed distribution with over 1,000 categories for instance segmentation~\cite{gupta2019lvis}. COCO-Stuff incorporates background (stuff) annotations for panoptic segmentation~\cite{caesar2018coco}, and RefCOCO enables referring expression-based object localization~\cite{kazemzadeh2014referitgame}. Other works improve \textit{vision-language alignment}, such as COCO-Caption for image captioning~\cite{chen2015microsoft} and COCO-Text for scene text detection~\cite{veit2016coco}. Additionally, CD-COCO, which applies image distortions to test robustness~\cite{beghdadi2023cd}, and COCONut, which unifies multiple segmentation tasks~\cite{deng2024coconut}, explore \textit{scene complexity}. However, these datasets do not manipulate contextual relationships. Our dataset uniquely reconstructs scene context by \textit{replacing original objects with out-of-context alternatives via inpainting}, creating a testbed for \textit{context modeling} and \textit{fake localization} that enables the learning of complex scene semantics.

\noindent\textbf{Context reasoning.}
Context is essential for understanding object relationships and finding anomalies in complex scenes~\cite{oliva2007role}.
Biederman et al.~\cite{biederman1982scene} identified relational principles such as support, probability, and size that help identify contextual inconsistencies, which prior works apply using co-occurrence and support relationships for out-of-context detection~\cite{choi2012context}.
Others have shown that slight context changes can cause significant errors in object detection, and Acharya et al.~\cite{acharya2022detecting} introduced a Graph Contextual Reasoning Network modeling co-occurrence and relative position to detect out-of-context objects.
Context is particularly important for object detection~\cite{dvornik2018modeling,bomatter2021pigs}; a comprehensive review is in~\cite{wang2023context}.
Ours is the first work to feature a mixture of in- and out-of-context objects in large-scale data. While context remains under-explored in fake detection, Large Vision Language Models (LVLMs)~\cite{deitke2024molmo, liu2023visual, liu2024llavanext, team2023gemini, jin2023unified} offer a human-aligned approach to context reasoning; we are the first to leverage them for context-based fake detection.

\noindent\textbf{Fake image generation and detection}.
Image manipulation has evolved from traditional object insertion \cite{farid2009image,bugeau2010comprehensive} to GAN-based synthesis \cite{goodfellow,isola2017image,brock2018large,karrastero,karras2020analyzing} and diffusion models \cite{ho2020denoising, sha2023fake}, including Stable Diffusion \cite{rombach2022high}, DALL-E \cite{ramesh2021zero}, and ControlNet \cite{Zhang_2023_ICCV}. Emerging techniques further tailor diffusion methods for object manipulation, inpainting, and harmonization \cite{chen2024anydoorzeroshotobjectlevelimage,zhang2023controlcomcontrollableimagecomposition, sun2025ousac}, 3D \cite{liu2023zero1to3}, and relighting \cite{he2024diffrelight, jin2024neuralgafferrelightingobject}. 
For detection, earlier works classify at image-level \cite{dang2018deep,mo2018fake}, while advanced methods localize manipulations: PSCC-Net \cite{liu2022pscc} uses spatio-channel correlation, CAT-Net \cite{kwon2022learning} exploits JPEG artifacts, ManTra-Net \cite{wu2019mantra} captures manipulation traces, and TruFor \cite{guillaro2023trufor} combines RGB and noise-sensitive fingerprints.
For fake image datasets, 
e.g., GenImage \cite{zhu2023genimage}, CIFAKE \cite{bird2023cifakeimageclassificationexplainable}, DE-FAKE \cite{sha2023fake}, they focus on fully-synthesized images. Diffusion-based inpainting datasets like COCOGlide \cite{guillaro2023trufor} and TGIF \cite{mareen2024tgif} utilize COCO images
but they have limited image quantity, object replacements constrained to the same category (thus lacking out-of-context objects), and no context labels.
Our dataset contains significant out-of-context fake objects in real scenes with rich characteristics.

\section{Building \datasetName}
\label{sec:building}
We propose \datasetName{}, a novel context-oriented inpainted objects dataset derived from the widely-used Common Objects in Context (COCO) Dataset~\cite{lin2014microsoft}. 
We specifically choose COCO as our foundation for several reasons.
First, we focus on common, everyday objects where contextual reasoning is most meaningful and widely applicable.
COCO's 80 classes span diverse supercategories covering frequent real-world categories such as animals, food, vehicles, and household items, making it the standard benchmark for context-rich scenes.
Second, COCO provides reliable instance masks and annotations essential for high-quality inpainting.
Expanding to rare or unusual objects would dilute the contextual focus and introduce annotation noise, as contextual expectations become less defined for uncommon categories.
Leveraging COCO's image- and object-level annotations, 
\datasetName{} enhances COCO by systematically replacing objects with diffusion-based inpainting, providing images with enriched contextual diversity. Our dataset comprises 97,722 successfully inpainted images derived from COCO2017, with each inpainted object categorized as in- or out-of-context with reasoning and accompanied by information about the original object and its replacement. Task-specific evaluation protocols are detailed in their respective sections.

For each COCO image, we (1) randomly pick an object from the image and inpaint a new object, (2) verify the new object was successfully inpainted with object detection, and (3) perform context reasoning with LVLMs to classify the new object as in- or out-of-context.

\subsection{Inpainting}

We use Stable Diffusion~\cite{rombach2022high} for inpainting.
Inpainting a new object requires an original image, an inpainting mask, and a prompt.
For each COCO image, we randomly select an object. The object's mask is slightly dilated and enclosed in a bounding box as the inpainting mask.
Dilation and bounding boxes reduce residual artifacts from the original objects. For replacement, we randomly select one of COCO's 80 categories as the text prompt.

The vanilla inpainting pipeline often struggles to inpaint small objects.
To overcome this, we crop the original image around an enlarged inpainting area, inpaint, and then scale the results back to the original size.
We use alpha blending to seamlessly merge the inpainted region with the original image. 
We found that for stronger inpainting models, such as Flux~\cite{labs2025flux1kontextflowmatching} and Qwen Image~\cite{wu2025qwenimagetechnicalreport}, they focus more on visual fidelity on in-context objects and 
tend to sacrifice the ability to generate out-of-context objects and provide less precise spatial control over the inpainting region.
More details on inpainting and the analysis of different inpainting models are in the supplementary materials.

\begin{figure}[t]
    \centering
    \includegraphics[height=2in]{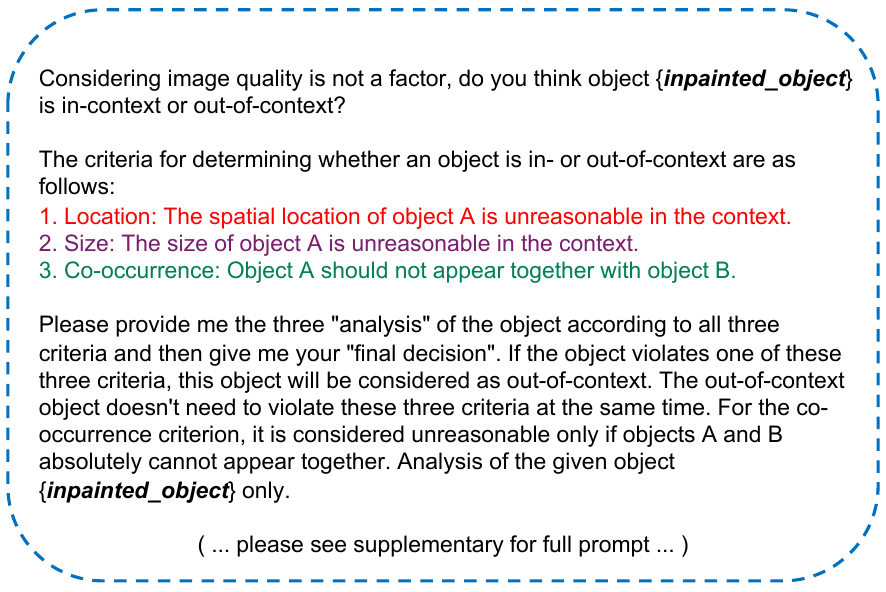}

    \caption{Context reasoning prompt for LVLMs.}
    \label{fig:context_reasoning}
\end{figure}


\subsection{Context reasoning}
\label{sec:context-reasoning}
We annotate context labels to all inpainted objects through a multi-model consensus process. 
Rather than relying on one single large model,
three state-of-the-art LVLMs (Molmo-72B~\cite{deitke2024molmo}, Qwen2.5-VL-72B~\cite{bai2025qwen2}, and InternVL3.5-38B~\cite{wang2025internvl3}) independently evaluate each inpainted image according to three context criteria: location, size, and co-occurrence.
We design a structured prompt (Figure~\ref{fig:context_reasoning}) to guide the LVLMs in performing this contextual reasoning.
For every image, each model first provides criterion-level reasoning by examining (1) whether the spatial placement of the object is appropriate for the scene layout, (2) whether its scale aligns with the scene geometry, and (3) whether its simultaneous presence with existing objects is contextually plausible. 
An object is labeled as \emph{out-of-context} if it violates any of these criteria; otherwise, it is deemed \emph{in-context}.

After generating criterion-level analyses, each LVLM produces a final binary judgment (in-context vs.\ out-of-context). 
To ensure the reliability of our context annotations, we retain labels only for images where 
a unanimous agreement is reached.
This consensus requirement yields 73,929 images for which we release both the final in-/out-of-context labels and the complete criterion-level reasoning from all LVLMs. 
The remaining images are still 
part of \datasetName{}, but we do not provide context annotations for them due to 
labeling ambiguity. 

\subsection{Comparison with prior datasets}
\label{sec:dataset_comparisons}


\begin{table}[t]
\renewcommand{\arraystretch}{1.2}
\setlength{\tabcolsep}{4pt}
\centering
\caption{\datasetName{} significantly enhances COCO in new ways.}
\label{tab:dataset_comparison}
\resizebox{\columnwidth}{!}{
\begin{tabular}{l|cccc}
\toprule
& \textbf{COCO (2017)} & \textbf{COCO-Stuff} & \textbf{COCOGlide} & \textbf{Ours} \\
\midrule
Dataset size & 123K & 164K & 512 & 97,722 \\
\hline
Unique source images & 123K & 164K & 512 & 97,722 \\
\hline
Replacement classes & - & - & same as original & 80 \\
\hline
Out-of-context images & - & - & - & \ding{52} \\
\hline
Context reasoning & - & \ding{52} & - & \ding{52} \\
\hline
Inpainting & - & - & \ding{52} & \ding{52} \\
\hline
Labels & objects & \makecell{objects, \\ scenes} & (fake) objects & \makecell{(fake) objects,\\ context reasoning} \\
\bottomrule
\end{tabular}
}
\end{table}


\datasetName{} extends COCO by introducing contextual diversity with inpainted objects (Table~\ref{tab:dataset_comparison}). Unlike COCO and COCO-Stuff, which primarily focus on object and scene (stuff) annotations, \datasetName{} provides intentional out-of-context scenarios, enabling research in contextual reasoning and fake localization. While COCOGlide \cite{guillaro2023trufor} also applies inpainting to COCO, it has significantly less images, and each replacement object has the same class as its original counterpart, thus the context is unaltered. 

\section{Manual Verification}
\label{sec:verification}
To validate the reliability of our automated models for object detection and context classification, we conducted a comprehensive manual verification study. We manually annotated 1,000 images to validate the successful rate of our final dataset, 
and assessed whether successfully inpainted objects were in- or out-of-context from human judgment.

\subsection{Verification on inpainted object detection}
\label{sec:verification-on-inpainted-object-detection}
To select an appropriate detector for our automated inpainting verification pipeline, we evaluated several COCO-trained models on 1,000 human-labeled images where annotators recorded whether the target object was successfully inpainted within the mask region. As shown in Table~\ref{tab:object_detection}, while GroundingDINO~\cite{liu2024grounding} achieved the highest precision (96.14\%), its low recall (38.66\%) would discard too many successful inpainting results. YOLOv8x~\cite{redmon2016you} achieved the best balance with an F1 score of 78.75\%, combining high precision (93.78\%) with reasonable recall (67.86\%). 
We therefore selected YOLOv8x as our detector
and use it to verify inpainted objects:
if YOLOv8x detects the target object within the inpainting mask, the process is deemed successful; for failed cases, we complete up to two additional rounds of inpainting and verification; images that fail all three rounds are discarded, resulting in 97,722 successful images. Notably, false positives often correspond to lower-quality inpaintings where the target object is present but poorly rendered, rather than complete inpainting failures.

\begin{table}[t]
\centering
\small
\begin{minipage}[t]{0.48\textwidth}
\centering
\caption{Object detection performance on inpainted objects.}
\label{tab:object_detection}
\begin{tabular}{@{}lcccc@{}}
\toprule
Metric & YOLOv8x & YOLOv11x & MMDet & GroundDINO \\
\midrule
Precision & 93.78 & 94.80 & 86.87 & 96.14 \\
Recall & 67.86 & 50.93 & 58.54 & 38.66 \\
F1 & \textbf{78.75} & 66.26 & 69.95 & 55.14 \\
\bottomrule
\end{tabular}
\end{minipage}
\end{table}

\subsection{Verification on context classification}
\label{sec:verification-on-context}
Following the procedure in Section~\ref{sec:context-reasoning}, we retained 73,929 images where all three LVLMs reached unanimous agreement on the final decision, with 64,879 labeled as out-of-context and 9,050 as in-context. To validate the quality of these context annotations, we randomly sampled 1,000 images (300 in-context, 700 out-of-context) for human verification. Human annotators classified each image using the same context criteria. The LVLMs consensus achieved 95.10\% agreement with human annotations, confirming the high reliability of our labels.

Among the 1,000 annotated images, we identified 164 cases where the three LVLMs agreed on the final out-of-context decision but disagreed on at least one specific criterion. Human evaluation of these disagreements revealed that Qwen2.5VL-72B achieved the highest accuracy at 87.80\% (144/164). For the remaining 836 images where all LVLMs agreed on both the final decision and all criteria, human annotators assessed reasoning quality. Qwen2.5VL-72B received 94.70\% preference for its detailed and interpretable explanations. Based on these validation results, we adopt Qwen2.5VL-72B's reasoning as the primary annotations for all 73,929 images in our dataset, with the consensus from MolMo and InternVL serving to ensure annotation reliability through multi-model verification.

\section{\datasetName-enabled Context Tasks}
\label{sec:analyses}
\begin{figure*}[h!]
    \centering
    \hspace*{0.5cm} 
    \renewcommand{\arraystretch}{1}
    \setlength{\tabcolsep}{.5pt}
    \begin{tabular}{>{\centering\arraybackslash}m{0.14\linewidth} 
                    >{\centering\arraybackslash}m{0.14\linewidth} 
                    >{\centering\arraybackslash}m{0.14\linewidth} 
                    >{\centering\arraybackslash}m{0.14\linewidth} 
                    >{\centering\arraybackslash}m{0.20\linewidth} 
                    >{\centering\arraybackslash}m{0.2\linewidth}}
        Original & Mask & Inpainted & Fake Det. & Context Reasoning & \footnotesize{Objects-from-Context} \\

        \includegraphics[width=\linewidth]{figures/fig6/test_paper_examples/originals/360.png} & 
        \includegraphics[width=\linewidth]{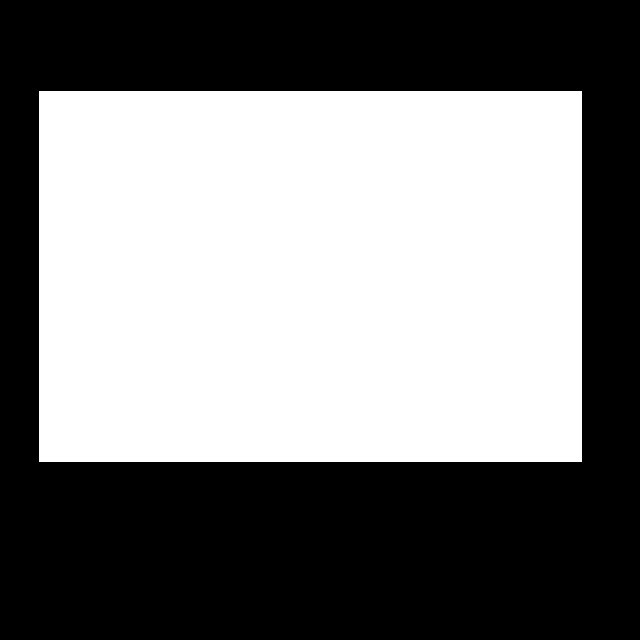} & 
        \includegraphics[width=\linewidth]{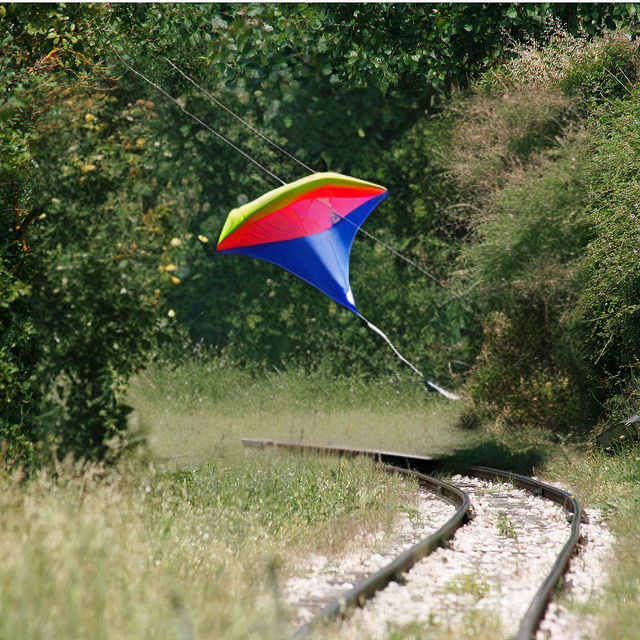} & 
        \includegraphics[width=\linewidth]{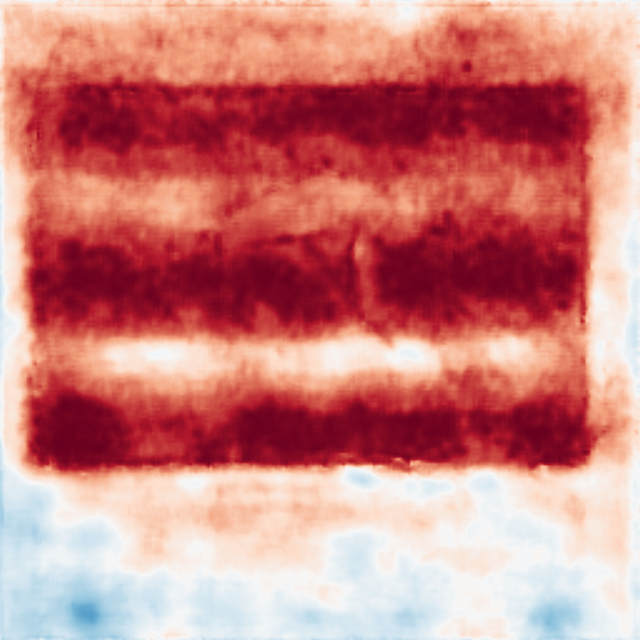} & 
        \scriptsize \textit{``A kite can reasonably be \textcolor{purple}{flying in the air above a railway track}... The kite appears to be \textcolor{violet}{appropriately sized for flying, neither too large nor too small for the environment.}''}\newline 
        Final decision: In-context. & 

        \includegraphics[width=0.8in]{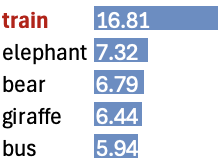}\\

        \includegraphics[width=\linewidth]{figures/fig6/test_paper_examples/originals/213.png} & 
        \includegraphics[width=\linewidth]{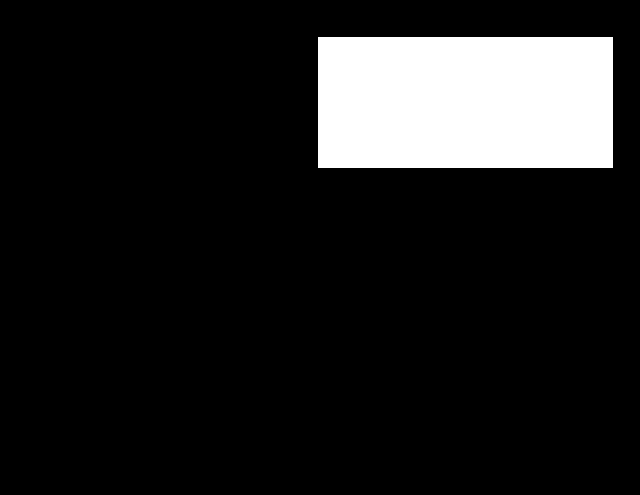} & 
        \includegraphics[width=\linewidth]{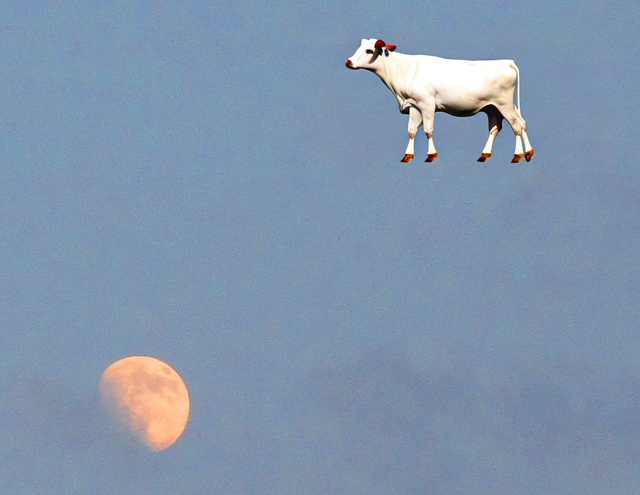} & 
        \includegraphics[width=\linewidth]{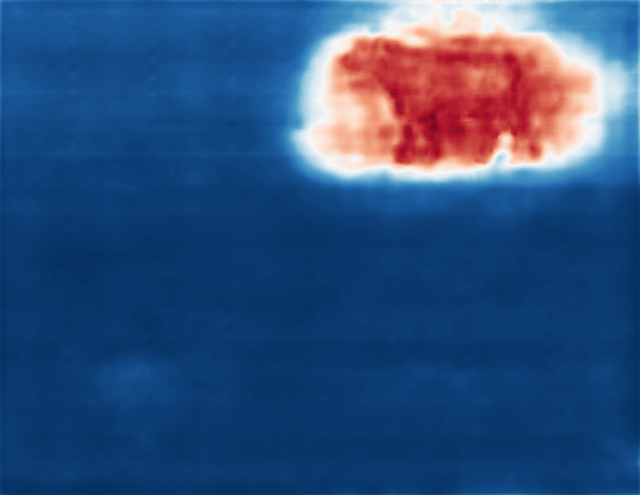} & 
        \scriptsize \textit{``\textcolor{purple}{A cow should not be in the sky.} It is \textcolor{teal}{impossible for a cow to appear together with the moon.}''}\newline 
        Final decision: Out-of-context. & 
        \includegraphics[width=0.8in]{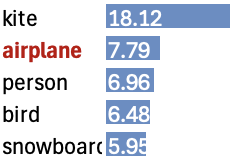}\\

        \includegraphics[width=\linewidth]{figures/fig6/test_paper_examples/originals/3005.png} & 
        \includegraphics[width=\linewidth]{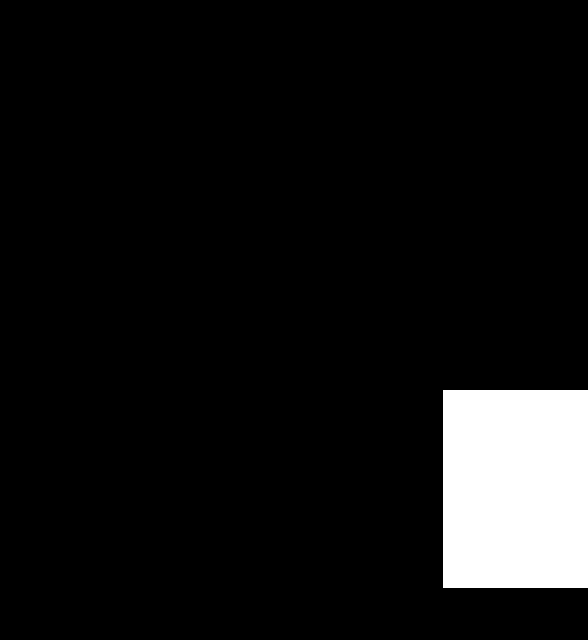} & 
        \includegraphics[width=\linewidth]{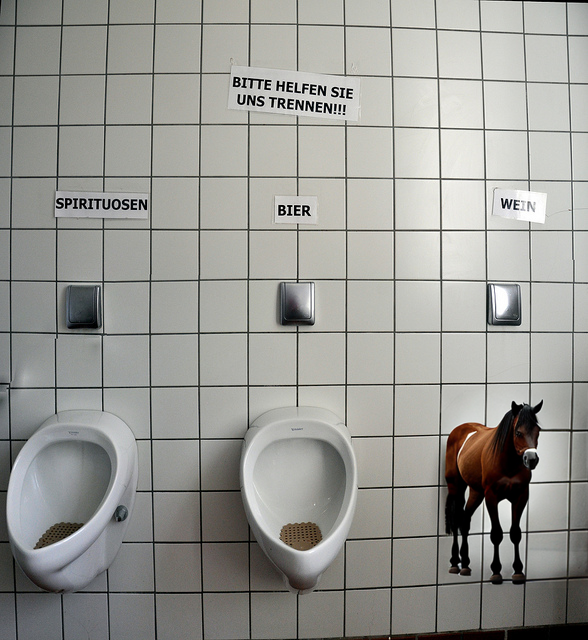} & 
        \includegraphics[width=\linewidth]{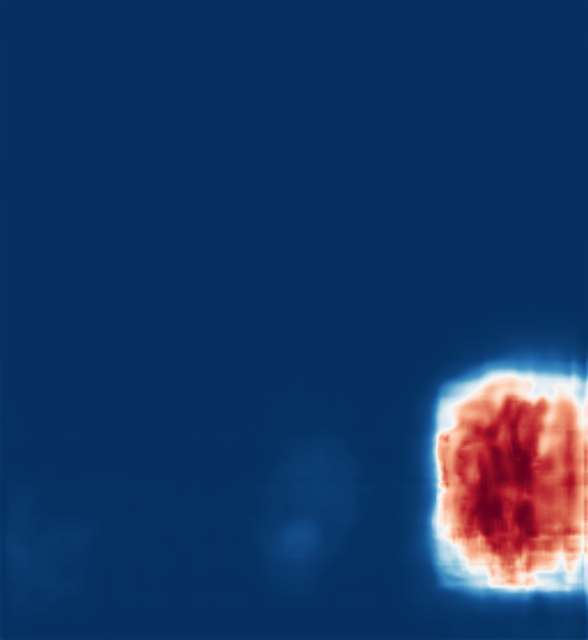} & 
        \scriptsize \textit{``A horse in a bathroom, especially next to urinals, is highly unusual and unreasonable... \textcolor{teal}{Horses and urinals are completely unrelated objects that should not appear together...}
''}\newline 
        Final decision: Out-of-context. & 
        \includegraphics[width=0.8in]{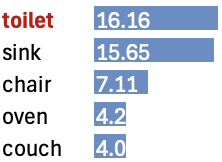}\\

        \includegraphics[width=\linewidth]{figures/fig6/test_paper_examples/originals/3617.png} & 
        \includegraphics[width=\linewidth]{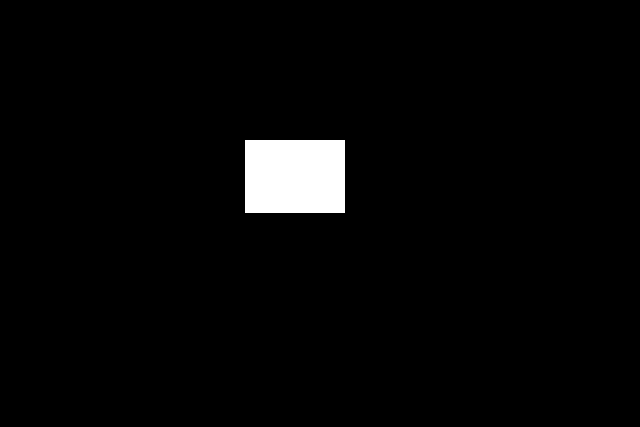} & 
        \includegraphics[width=\linewidth]{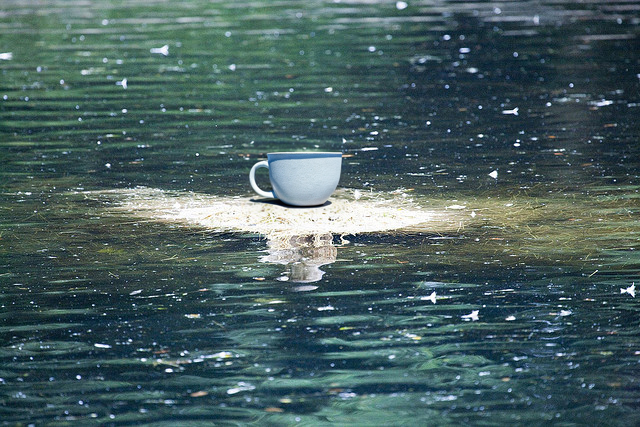} & 
        \includegraphics[width=\linewidth]{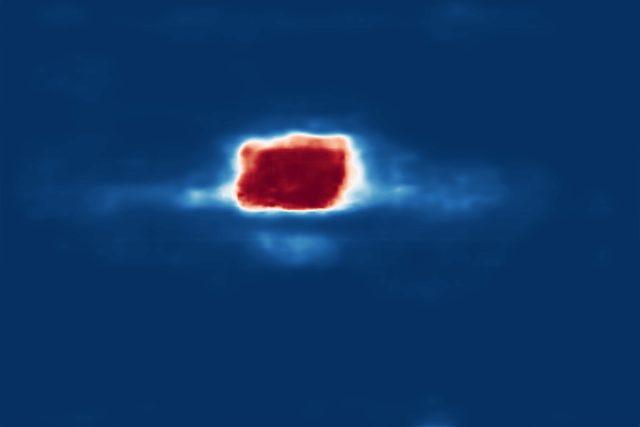} & 
        \scriptsize \textit{''Cups and water are \textcolor{teal}{common together}, but a cup floating on its own \textcolor{purple}{in a natural body of water} is not reasonable}.''\newline 
        Final decision: Out-of-context. & 
        \includegraphics[width=0.8in]{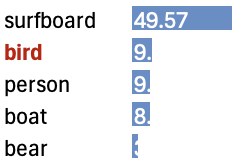}\\

        
        \includegraphics[width=\linewidth]{figures/fig6/test_paper_examples/originals/4188.png} & 
        \includegraphics[width=\linewidth]{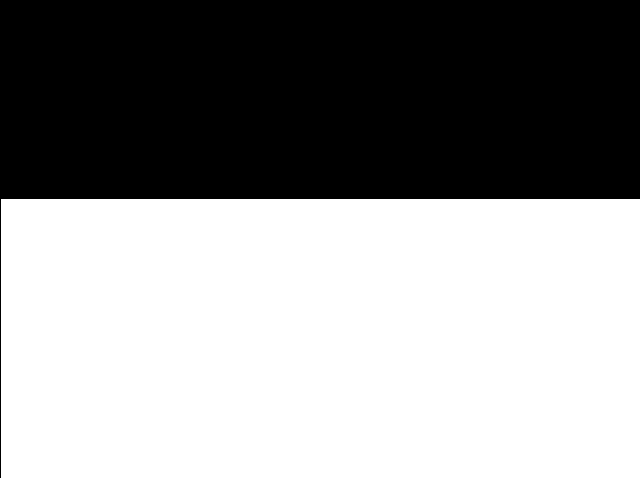} & 
        \includegraphics[width=\linewidth]{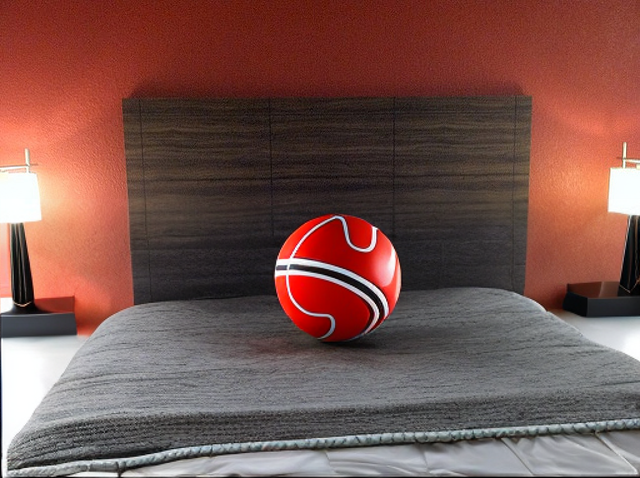} & 
        \includegraphics[width=\linewidth]{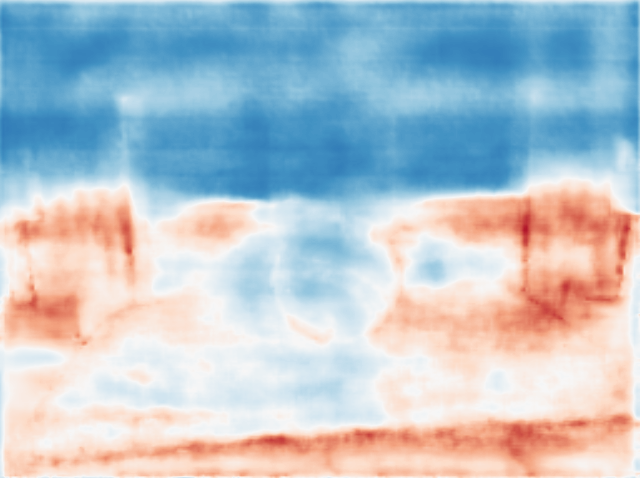} & 
        \scriptsize \textit{``A soccer ball is reasonable in a bedroom setting...\textcolor{teal}{Soccer balls can appear together with beds, as they are both common items in a bedroom.}''}\newline 
        Final decision: In-context. & 

        \includegraphics[width=0.8in]{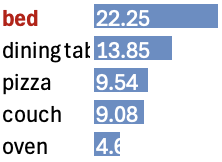}\\

        \includegraphics[width=\linewidth]{figures/fig6/test_paper_examples/originals/549.png} & 
        \includegraphics[width=\linewidth]{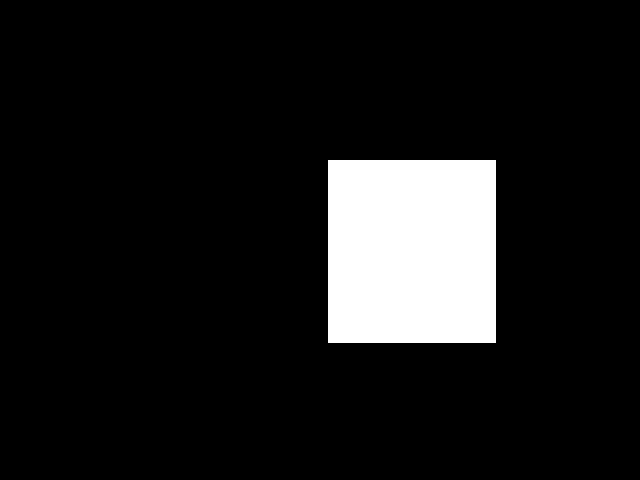} & 
        \includegraphics[width=\linewidth]{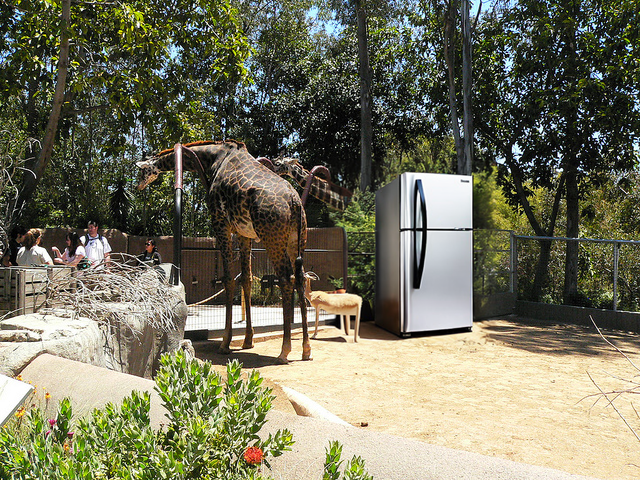} & 
        \includegraphics[width=\linewidth]{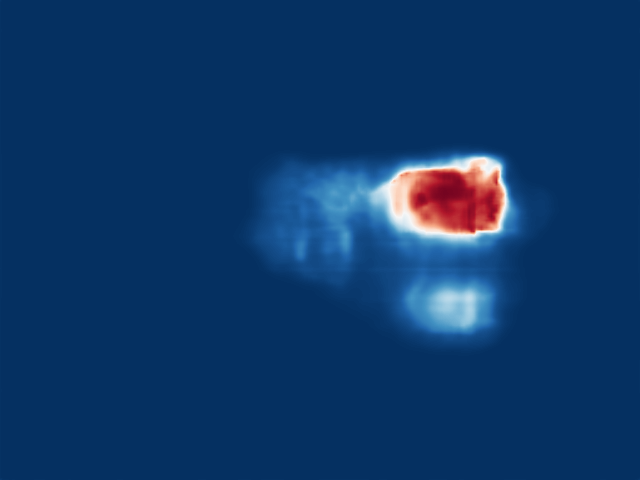} & 
        \scriptsize \textit{``A refrigerator is \textcolor{teal}{highly unusual and out of place in a zoo enclosure}, especially one with giraffes and other animals...}\newline 
        Final decision: Out-of-context.'' & 

        \includegraphics[width=0.8in]{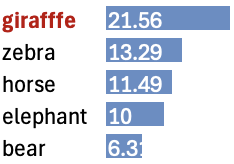}\\

    \end{tabular}
    \caption{Inpainting, fake detection, and objects-from-context results. 
    Context reasoning responses are color-coded by \textcolor{purple}{location}, \textcolor{violet}{size}, and \textcolor{teal}{co-occurrence}.
    Original objects are in \textcolor{BrickRed}{red}. Inpainted objects: kite, cow, horse, cup, sports ball, refrigerator.
    }
\label{fig:examples_of_pipeline}
\end{figure*}

With \datasetName{} as our testbed, we explore the role of context in several tasks. 
We demonstrate how our data can be used to train efficient context classifiers through knowledge distillation, fine-tuning smaller models to evaluate objects based on our three context criteria.
We then propose a novel Objects-from-Context task for predicting which objects naturally belong in a given scene.
Finally, we evaluate state-of-the-art fake detection models and show that context can enhance fake localization without any fine-tuning.
These applications demonstrate the versatility of \datasetName{} and underscore the fundamental role of context in visual understanding. 
Examples of those tasks are in
Figure~\ref{fig:examples_of_pipeline}.


\subsection{Fine-grained context classification}
\label{sec:fine-grained-context}
Our context classification task addresses a fundamental question: \textit{``Does an object belong in this context?''} To answer this question effectively at scale, we need models that can perform accurate context reasoning while being practical for deployment. While our verification showed that Qwen2.5-VL-72B achieved excellent performance in context classification, its large size makes it impractical for many real-world deployment scenarios.

To address this challenge, we employ knowledge distillation~\cite{liu2023visual,hsieh2023distilling, yang2025concept, alpaca}, a proven technique that transfers knowledge from large teacher models to smaller, more efficient student models. We use Qwen2.5-VL-72B as the teacher to distill its knowledge into three independent Qwen2.5-VL-3B student models—one for location, one for size, and one for co-occurrence reasoning. This distillation process reduces the model size by 24× while maintaining most of the reasoning capabilities, making context classification feasible for practical deployment.
Note that this knowledge distillation task is possible thanks to the rich context reasoning labels in \datasetName.

We use the 73,929 unanimous consensus images from \datasetName{} to create balanced evaluation splits for training our distilled models. Specifically, we sample 24,000 images maintaining a 1:1 ratio of in-context to out-of-context examples, which we divide into 20,000 for training, 2,000 for validation, and 2,000 for testing. This balanced sampling ensures unbiased evaluation of context reasoning capabilities.
We use context reasoning responses from QwenVL as the ground truth (discussed in Section~\ref{sec:verification-on-context}). Importantly, each student model learns not only the final decision (in-context or out-of-context) but also the reasoning process for its specific criterion, enabling interpretable explanations.
\begin{table}[t]
\centering
\caption{Context reasoning accuracy (\%) on COinCO test set and original COCO images. Results on original COCO confirm models learn genuine context reasoning
rather than inpainting artifacts.}
\label{tab:context_prediction}
\footnotesize
\begin{tabular}{lcccccc}
\toprule
& \multicolumn{3}{c}{COinCO Test Set}  & \multicolumn{3}{c}{Original COCO Images} \\
\cmidrule(lr){2-4} \cmidrule(lr){5-7}
Criterion & Before & After & $\Delta$ & Before & After & $\Delta$ \\
\midrule
Size & 55.4 & 65.9 & +10.5 & 52.6 & 83.2 & +30.6 \\
Location & 67.6 & 75.4 & +7.8 & 47.6 & 91.4 & +43.8 \\
Co-occurrence & 75.5 & 79.9 & +4.4 & 89.0 & 87.0 & -2.0 \\
\midrule
Average & 66.2 & 73.7 & +7.6 & 63.1 & 87.2 & +24.1 \\
\bottomrule
\end{tabular}
\vspace{-3mm}
\end{table}

Table~\ref{tab:context_prediction} shows consistent improvements for each student model across all criteria. On the COinCO test set, size reasoning improves most (+10.5\% accuracy), while co-occurrence, despite the strongest baseline, reaches nearly 80\% accuracy, confirming successful knowledge transfer from the 72B teacher to 3B students.

\noindent\textbf{Validation on original COCO images.} To verify that our models learn genuine semantic relationships and context rather than fitting to inpainting artifacts, we test them on 2,000 original COCO images without any manipulation. 
The ground-truth values of all three criteria are true as the objects are considered as in-context.
The co-occurrence model achieves 89.0\% accuracy on original images before fine-tuning, demonstrating it captured natural object relationships. After fine-tuning on COinCO, accuracy slightly decreases to 87.0\%, suggesting the model already learned robust co-occurrence patterns that need no further adjustment. The location model shows the largest improvement after fine-tuning (+43.8\% accuracy), indicating that exposure to diverse object placements in COinCO helps the model learn spatial reasoning that transfers to real images. The size model also improves substantially (+30.6\% accuracy), learning appropriate scale relationships. For location and size reasoning, models achieve an average improvement of +37.2\% accuracy on original COCO images, with the high performance on unmanipulated images confirming these models understand genuine contextual principles rather than memorizing dataset-specific patterns.

Each model provides interpretable natural language explanations for its decisions. Through distillation, these 3B models achieve practical inference speeds while maintaining reasoning quality~\cite{hsieh2023distilling,xu2024surveyknowledgedistillationlarge}. The modular design allows criterion-specific refinements, demonstrating that complex contextual understanding can be decomposed into focused reasoning tasks. While the 72B teacher model is too slow for practical deployment, these efficient 3B student models make context reasoning feasible for real-world applications.

\begin{figure*}[t]
    \centering
    \small
    
    
    
    \begin{tabular}{@{}c@{\hspace{10pt}}c@{\hspace{10pt}}c@{}}
        \includegraphics[width=0.30\textwidth]{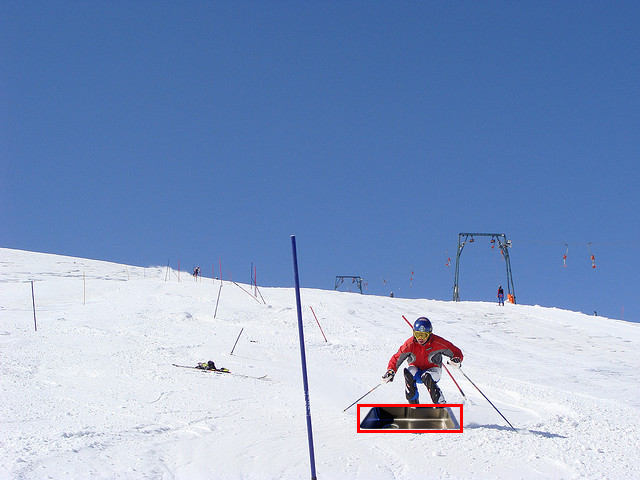} &
        \includegraphics[width=0.30\textwidth]{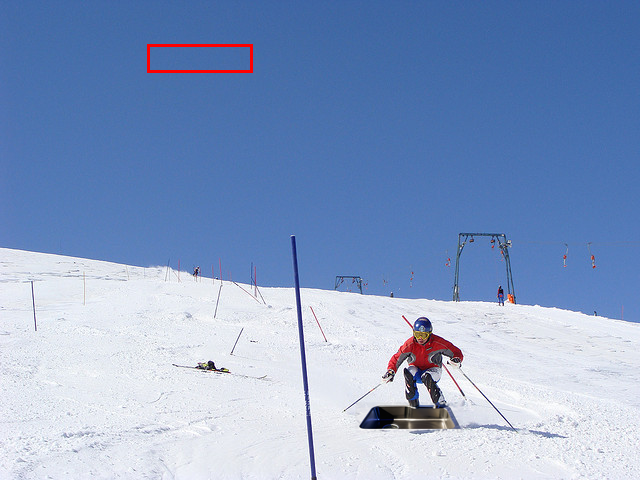} &
        \includegraphics[width=0.30\textwidth]{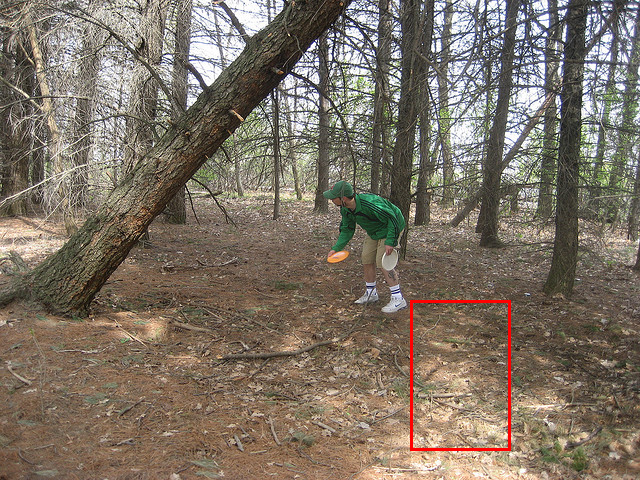}
    \end{tabular}
    
    \vspace{8pt}
    
    \begin{tabular}{@{}c@{\hspace{10pt}}c@{\hspace{10pt}}c@{}}
        \begin{tabular}{lr@{\hspace{8pt}}|@{\hspace{8pt}}lr}
        \multicolumn{2}{c@{\hspace{8pt}}|@{\hspace{8pt}}}{\textbf{Instance}} & 
        \multicolumn{2}{c}{\textbf{Clique}} \\
        \hline
        \textcolor{red}{\textbf{skis}} & \textbf{31.6} & sports & 79.2 \\
        snowboard & 24.3 & accessory & 6.6 \\
        kite & 14.8 & person & 4.9 \\
        surfboard & 7.0 & vehicle & 4.2 \\
        backpack & 5.9 & animal & 3.0
        \end{tabular} &
        
        \begin{tabular}{lr@{\hspace{8pt}}|@{\hspace{8pt}}lr}
        \multicolumn{2}{c@{\hspace{8pt}}|@{\hspace{8pt}}}{\textbf{Instance}} & 
        \multicolumn{2}{c}{\textbf{Clique}} \\
        \hline
        \textcolor{red}{\textbf{kite}} & \textbf{22.9} & sports & 75.1 \\
        skis & 22.1 & accessory & 6.1 \\
        snowboard & 19.6 & vehicle & 5.7 \\
        surfboard & 8.6 & person & 5.5 \\
        person & 5.5 & animal & 4.2
        \end{tabular} &
        
        \begin{tabular}{lr@{\hspace{8pt}}|@{\hspace{8pt}}lr}
        \multicolumn{2}{c@{\hspace{8pt}}|@{\hspace{8pt}}}{\textbf{Instance}} & 
        \multicolumn{2}{c}{\textbf{Clique}} \\
        \hline
        \textcolor{red}{\textbf{horse}} & \textbf{5.0} & animal & 37.2 \\
        zebra & 5.0 & vehicle & 14.9 \\
        giraffe & 4.9 & sports & 12.1 \\
        person & 4.6 & outdoor & 10.7 \\
        elephant & 4.4 & accessory & 6.8
        \end{tabular}
    \end{tabular}
    
    \caption{Object-from-Context prediction. A red box in each image indicates the query region. The top row shows three examples (two inpainted, one original COCO). The bottom row lists instance-level and clique-level predictions with their probabilities (P(\%)). Objects in \textcolor{red}{red} are the top predictions.}
    \label{fig:object_from_context}
    \vspace{-0.2cm}
\end{figure*}

\subsection{Objects-from-Context prediction}
\label{sec:original-object}

While context classification models determine if an existing object belongs in a scene, our Objects-from-Context prediction tackles a different question: \textit{``What object(s) fit this context?''} This task tests a model's contextual comprehension of real images 
and aligns with the Context Challenge in \cite{torralba2003contextual}. We formulate this task at two levels:
1) \textbf{Instance-level prediction}. Predicting the object class among COCO's 80 classes.
2) \textbf{Clique-level prediction}. Predicting the clique an object belongs to, using COCO's 
super-categories: accessory, animal, appliance, electronic, food, furniture, indoor, kitchen, outdoor, person, sports, and vehicle.

We design a model that takes two inputs: the inpainted image and a binary mask indicating the target region. During training, we use the dilated bounding box of the original object as the mask and its class as the label. We use the VAE encoder from Stable Diffusion to extract latents from both inputs, which are processed by an MLP that predicts across COCO's 80 classes. For clique-level evaluation, we map predicted object classes to their corresponding super-categories and check if they match the original objects. 
Importantly, our model can handle arbitrary mask sizes and locations during inference, making it adaptable for broader context reasoning applications. Detailed model architectures and training information are in the supplementary material.
For comparison, we implement a baseline that ranks candidate objects based on the co-occurrence frequency of other objects in a scene~\cite{mack2011object}. 
If no other objects are present, this baseline defaults to random selection. We evaluate our Objects-from-Context model on the 2,402 images inpainted from the COCO2017 validation set.

\begin{table}[t]
\centering
\caption{
Instance-level and clique-level object-from-context prediction accuracy.
}
\label{tb:context-from-object-accuracy}
\small
\resizebox{0.99\linewidth}{!}{
\begin{tabular}{lcccccc}
\toprule
\multirow{2}{*}{Method} & \multicolumn{3}{c}{Instance-level (\%)} & \multicolumn{3}{c}{Clique-level (\%)} \\
\cmidrule(lr){2-4} \cmidrule(lr){5-7}
 & Top-1 & Top-3 & Top-5 & Top-1 & Top-3 & Top-5 \\
\midrule
Random & 1.25 & 3.75 & 6.25 & 8.33 & 25.00 & 41.67 \\
Co-occurrence~\cite{mack2011object} & 1.54 & 4.70 & 7.29 & 9.37 & 30.72 & 52.91 \\
\textbf{Ours} & \textbf{16.32} & \textbf{31.89} & \textbf{42.80} & \textbf{35.10} & \textbf{61.41} & \textbf{78.31} \\
\bottomrule
\end{tabular}
}
\vspace{-5pt}
\end{table}

Table~\ref{tb:context-from-object-accuracy} shows our model's superior grasp of contextual features, significantly outperforming random and co-occurrence baselines. At the instance level, the substantial improvement in top-3 and top-5 accuracy (31.89\% and 42.80\%) indicates that even when the model's top prediction is incorrect, the true object is often ranked among the most probable candidates. 
Our model demonstrates stronger performance at the clique level, correctly predicting semantically similar objects.
This result
demonstrates that our approach effectively learns the relationship between scene contexts and the objects that belong within them. Our model offers three key applications as shown in  Figure~\ref{fig:object_from_context}: (1) analyzing suspicious regions to predict what objects existed before manipulation, serving as a reference for anomaly detection and forensic analysis~\cite{pang2021deep}; (2) suggesting contextually appropriate objects based on spatial location within a scene~\cite{sun2020hidden,sun2017seeing}, enabling intelligent image editing~\cite{chen2024anydoorzeroshotobjectlevelimage}; and (3) versatile analysis of any image (including unmodified images) to identify contextually coherent objects in regions of interest, supporting context-aware content generation. This versatility advances context-driven visual understanding across multiple practical applications.

\subsection{Context-empowered fake localization}

\begin{table}[t]
  \centering
  \small
  \caption{SOTA fake detection performance on \datasetName.}
  \label{tab:pixel_detection}
  \begin{tabular}{lcccc}
  \toprule
  Method & Acc & F1 & AUC & AP \\
  \midrule
  ManTraNet~\cite{wu2019mantra} & 85.4 & 30.7 & 84.9 & 50.7 \\
  Trufor~\cite{guillaro2023trufor} & 89.7 & 55.9 & 93.4 & 73.6 \\
  PSCC-Net~\cite{liu2022pscc} & 89.5 & 46.8 & 95.4 & 79.5 \\
  \textbf{CAT-Net}~\cite{kwon2022learning} & \textbf{92.7} & \textbf{76.5} & \textbf{97.4} & \textbf{90.3} \\
  \bottomrule
  \end{tabular}
\end{table}

\begin{figure}[t]
  \centering
  \includegraphics[width=0.98\linewidth]{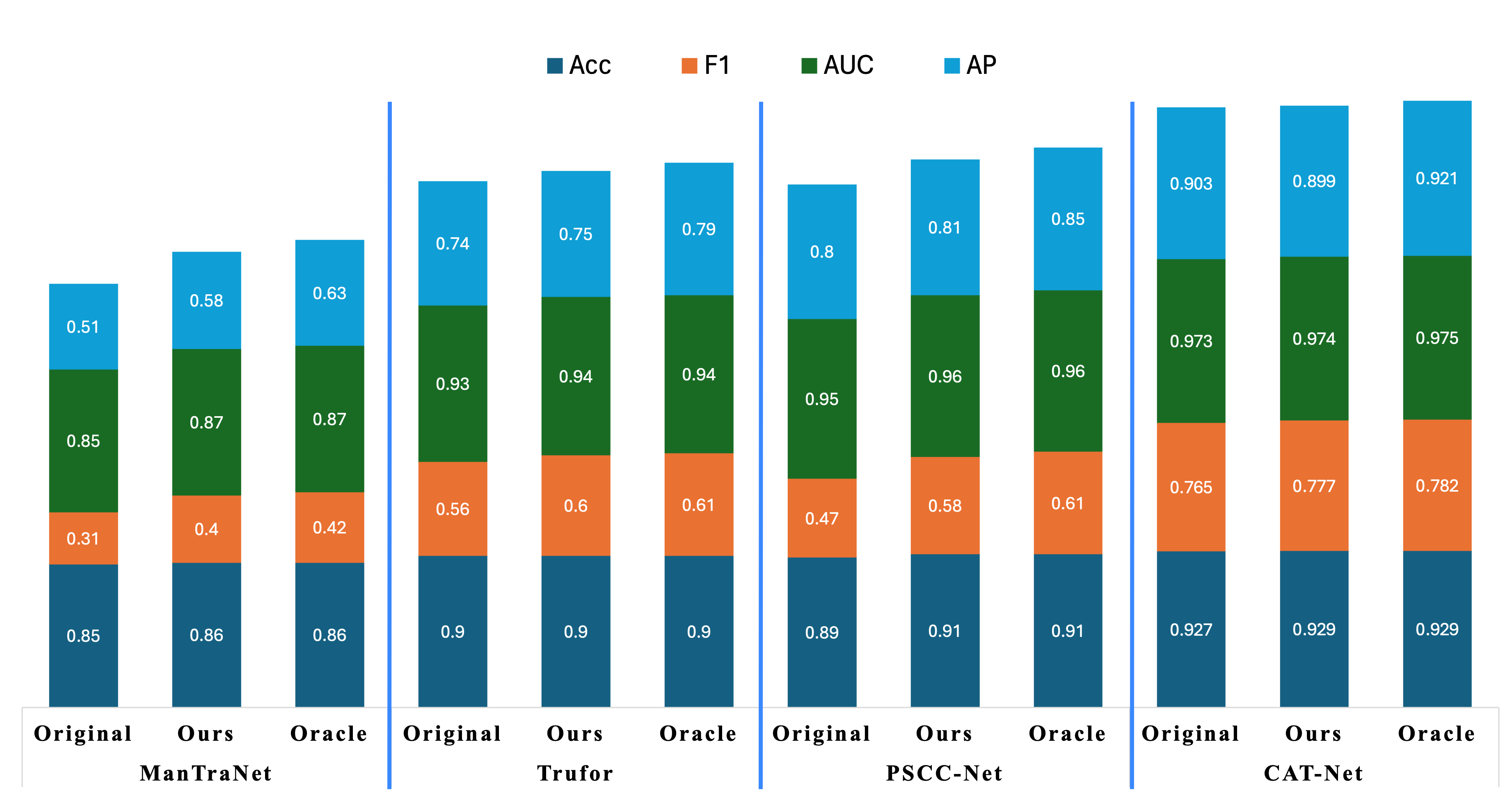}
  \caption{Context enhancement results.}
  \label{fig:context_enhancement}
\end{figure}
Fake localization aims to identify the specific synthetic regions in an image, producing a pixel-level map that is fine-grained compared to binary image-level classification.
In \datasetName{}, the ground truth mask for fake localization is the original object's entire bounding box used in inpainting.
We evaluate localization performance on the 2,402 test images derived from COCO2017 validation using four common pixel-level metrics~\cite{kwon2022learning, liu2022pscc}: F1 Score and Accuracy (threshold of $0.5$), Area Under Curve (AUC), and Average Precision (AP).

\begin{figure}[t]
    \centering
    \footnotesize 
    \renewcommand{\arraystretch}{1.0} 
    \setlength{\tabcolsep}{2pt} 
    \begin{tabular}{@{}cccccc@{}}
        \textbf{Image} & 
        \textbf{GT} & 
        \textbf{\cite{kwon2022learning}} & 
        \textbf{\cite{wu2019mantra}} & 
        \textbf{\cite{guillaro2023trufor}} & 
        \textbf{\cite{liu2022pscc}} \\
        \includegraphics[width=0.15\linewidth]{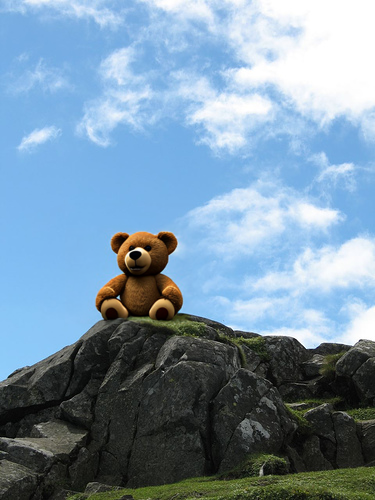} &
        \includegraphics[width=0.15\linewidth]{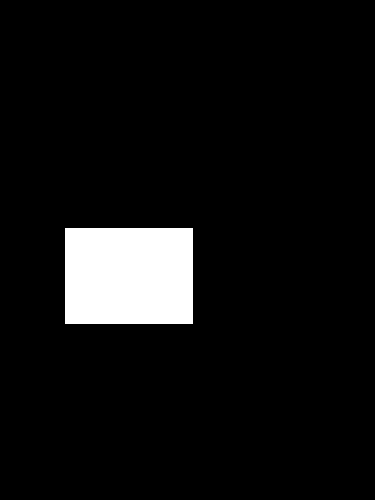} &
        \includegraphics[width=0.15\linewidth]{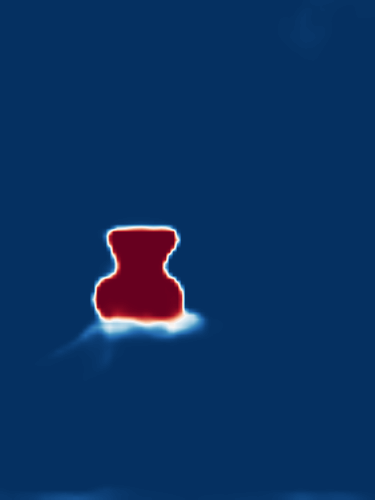} &
        \includegraphics[width=0.15\linewidth]{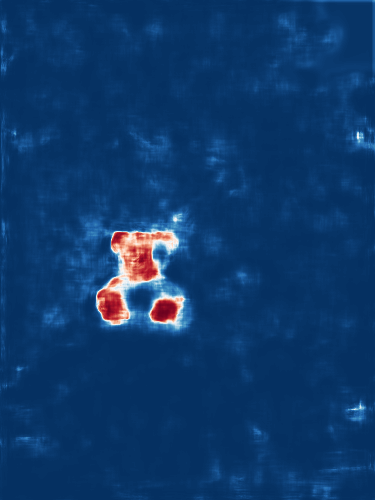} &
        \includegraphics[width=0.15\linewidth]{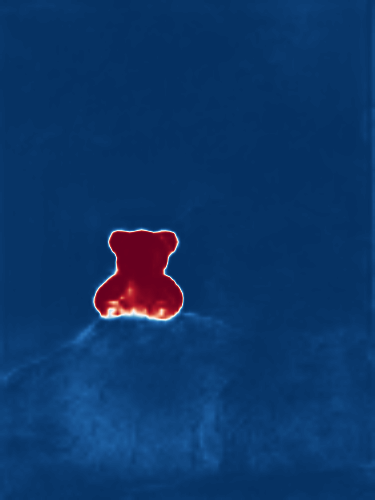} &
        \includegraphics[width=0.15\linewidth]{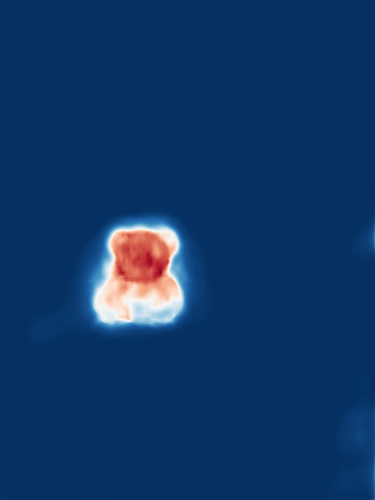} \\
        \includegraphics[width=0.15\linewidth]{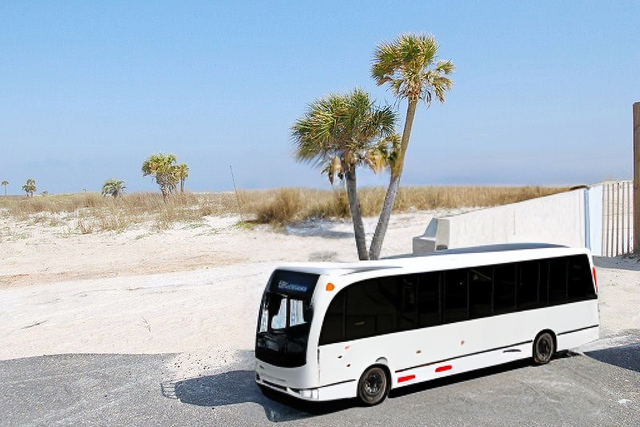} &
        \includegraphics[width=0.15\linewidth]{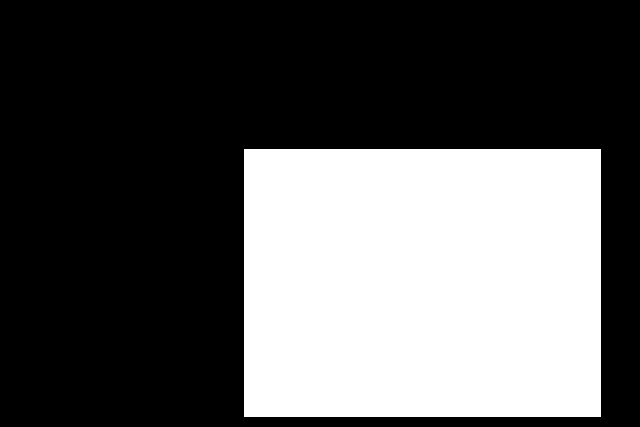} &
        \includegraphics[width=0.15\linewidth]{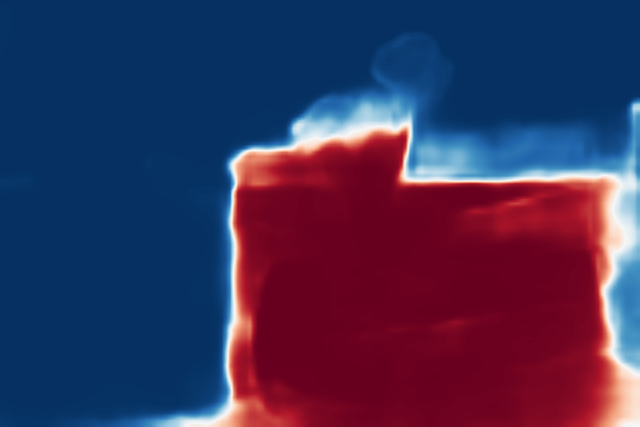} &
        \includegraphics[width=0.15\linewidth]{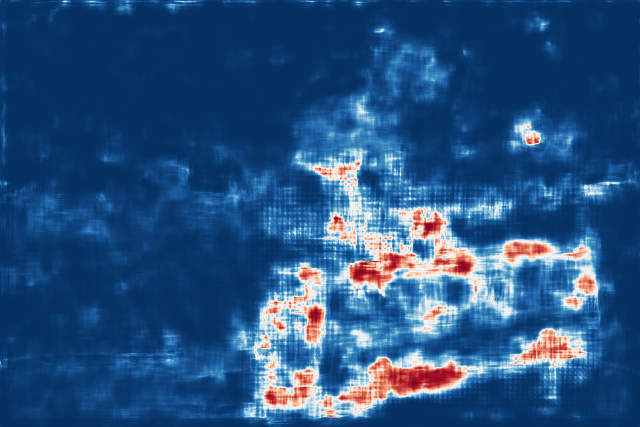} &
        \includegraphics[width=0.15\linewidth]{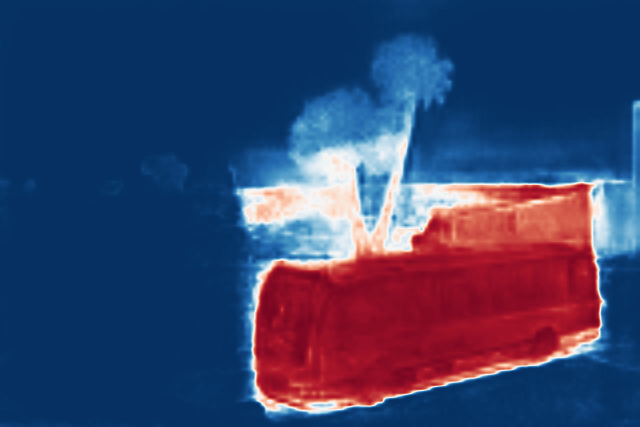} &
        \includegraphics[width=0.15\linewidth]{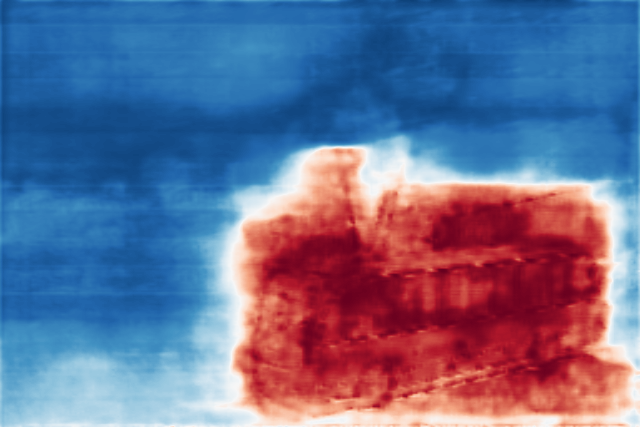} \\
        \includegraphics[width=0.15\linewidth]{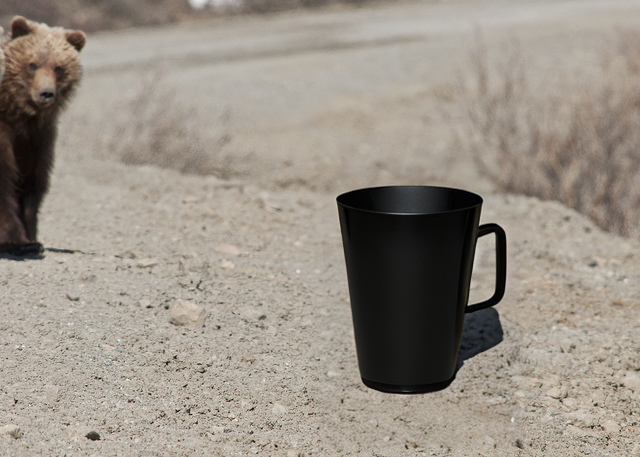} &
        \includegraphics[width=0.15\linewidth]{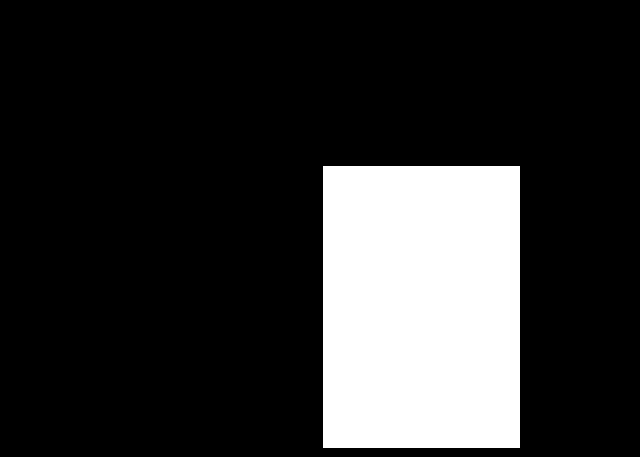} &
        \includegraphics[width=0.15\linewidth]{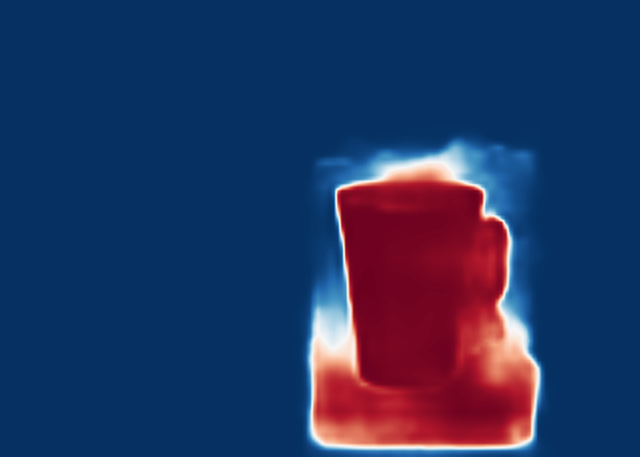} &
        \includegraphics[width=0.15\linewidth]{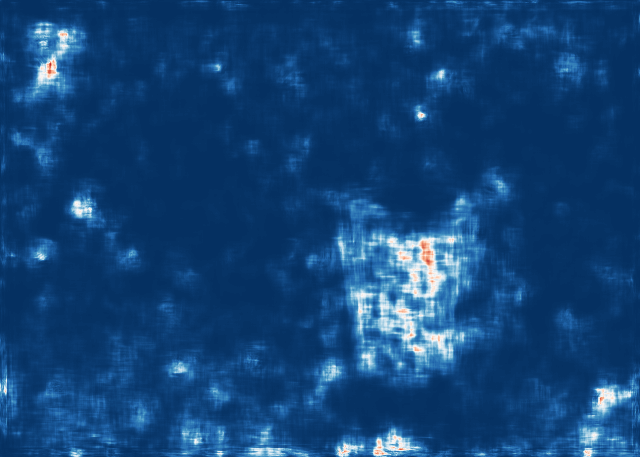} &
        \includegraphics[width=0.15\linewidth]{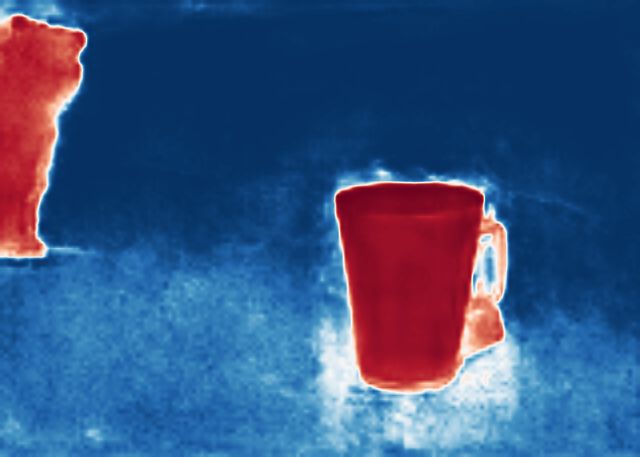} &
        \includegraphics[width=0.15\linewidth]{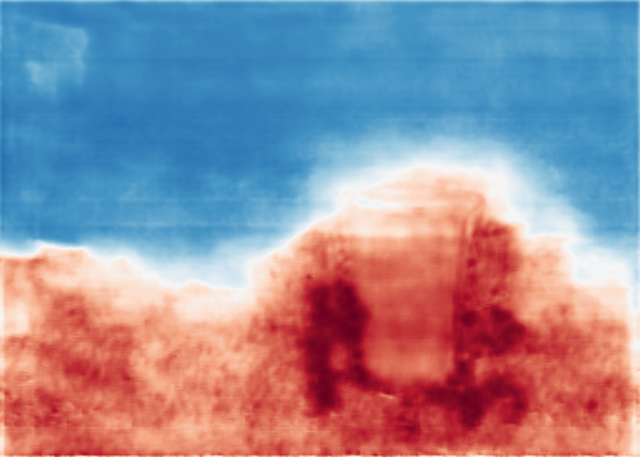} \\
        \includegraphics[width=0.15\linewidth]{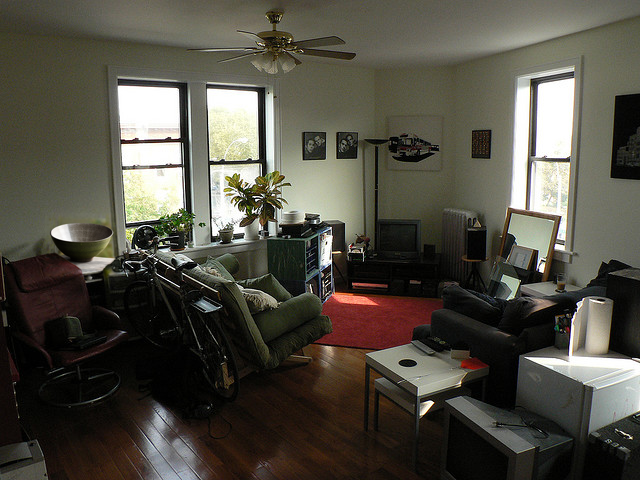} &
        \includegraphics[width=0.15\linewidth]{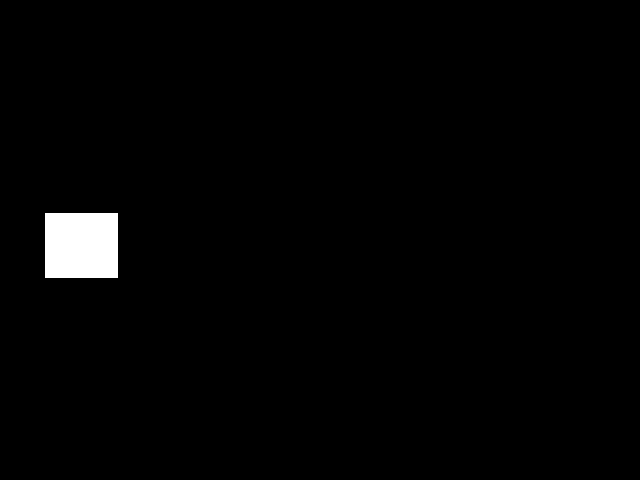} &
        \includegraphics[width=0.15\linewidth]{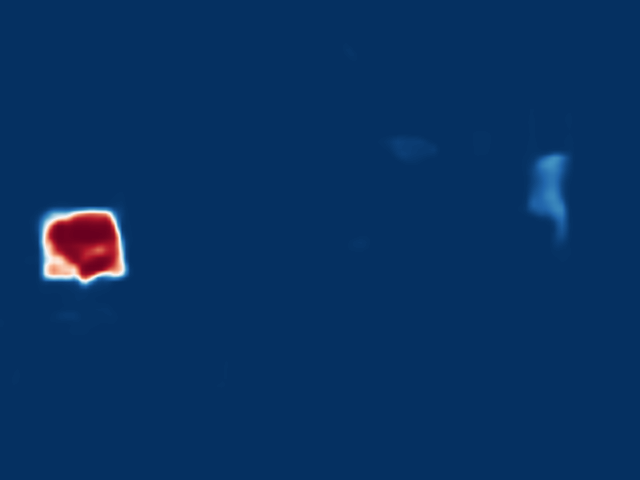} &
        \includegraphics[width=0.15\linewidth]{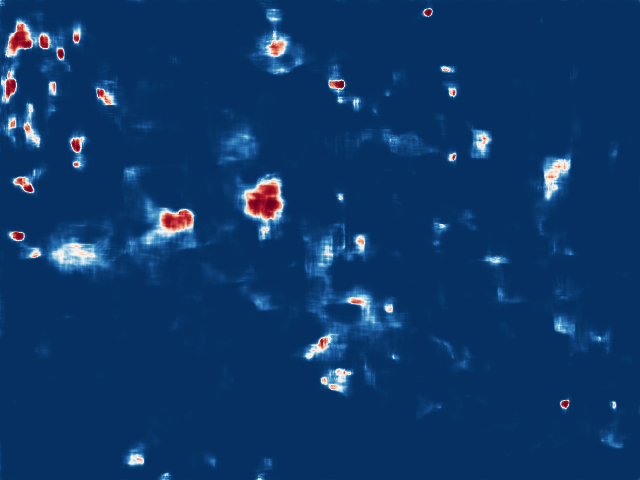} &
        \includegraphics[width=0.15\linewidth]{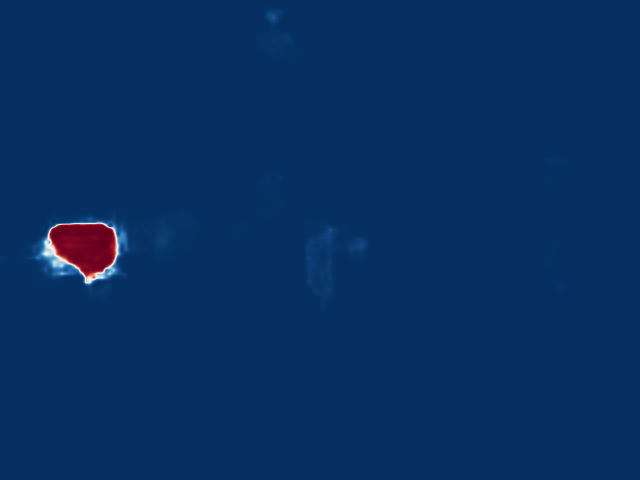} &
        \includegraphics[width=0.15\linewidth]{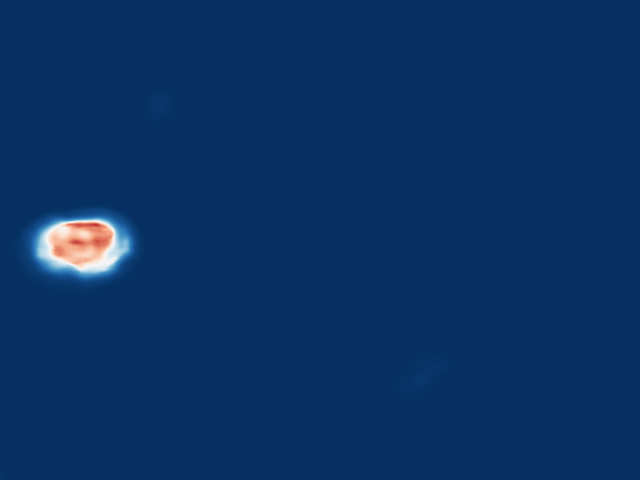} \\
    \end{tabular}
    \caption{Fake localization performance.}
    \label{fig:sota_results}
\end{figure}

\noindent\textbf{State-of-the-art performance.} 
We benchmark several SOTA fake detection
models on \datasetName{} (Table~\ref{tab:pixel_detection}).
The high accuracy and AUC scores ($>84\%$ for all models) can be attributed to the small average size of inpainted regions in our data. As most images are majority authentic, these metrics are biased towards high values. F1 and AP, which better reflect the precision of fake localization, reveal a clear performance gap among models. CAT-Net achieves the best performance across all metrics with $92.65\%$ accuracy, $76.47\%$ F1 score, $97.35\%$ AUC, and $90.33\%$ AP. Trufor obtains the second-best F1 score of $55.92\%$, while PSCC-Net shows strong performance in AUC ($95.38\%$) and AP ($79.52\%$). ManTraNet demonstrates relatively lower performance in F1 score ($30.68\%$) and AP ($50.72\%$). These findings highlight significant differences and room for improvement among SOTA fake localizers. Examples of these SOTA models’ predictions are shown in Figure~\ref{fig:sota_results}.


\noindent\textbf{Incorporating context in fake localization.}
To leverage contextual information for improved fake localization, we propose a simple yet effective way of enhancing the prediction scores of fake localization models in regions corresponding to out-of-context objects. Concretely, the context-enhanced prediction score $P'(x,y)$ at pixel $(x,y)$ is defined as $\min(P(x,y) \times \gamma,\, 1.0), \text{if } (x,y) \in M_{\text{OC}}$; $\text{otherwise as} P(x,y)$,
the original predicted fake score, $M_{\text{OC}}$ is the mask region of out-of-context object (predicted by context reasoning), $\gamma$ is a context enhancement factor, and the $\min$ function ensures the enhanced score $\leq 1.0$.

We evaluate this context-enhancement approach under two settings. In the \textbf{oracle} setting, we use ground truth annotations to enhance predictions within the fake object's mask region when it is out-of-context. This serves as an upper bound for performance gains.
In the \textbf{practical} setting, the fake objects are unknown. We propose the use of Molmo-7B~\cite{deitke2024molmo} to identify suspicious objects based on \textit{size} and \textit{location}, excluding \textit{co-occurrence} to prevent false positives. For instance, if an image contains only two objects, e.g., an apple and an inpainted traffic light, either might be seen as out-of-context. By focusing on physical violations, our method remains robust while effectively detecting fake objects. As Molmo may identify multiple out-of-context objects in an image, we define $M_{\text{OC}}$ as the union of these objects' masks. In both settings, we set the enhancement factor $\gamma$ to $5$ 
to enhance prediction scores in out-of-context regions. 
The analysis of $\gamma$ is in supplementary materials.

Figure~\ref{fig:context_enhancement} shows that our context-enhancement improves fake localization for SOTA detectors \textit{without any finetuning}.
The effectiveness of our method stems from the complementary nature of context detection and fake localization. Thus, even if LVLM 
mistakes
authentic objects as out-of-context, enhancement only strengthens the base model’s predictions.
Notably, out-of-context objects are not rare in real-world forgeries, especially in partial manipulation scenarios, where inserted objects often exhibit semantic inconsistencies in size, location, or scene compatibility. Our results highlight that context is valuable for fake detection.

\section{Conclusion}
\label{sec:conclusion}
We present \datasetName{}, a large-scale dataset featuring both in-context and out-of-context objects created through controlled diffusion-based inpainting. By strategically replacing COCO objects, \datasetName{} introduces diverse contextual configurations and provides rich criterion-level annotations generated through multiple LVLMs consensus. These annotations enable training efficient, interpretable context classifiers through knowledge distillation, and support a new Objects-from-Context prediction task that models what objects naturally belong in a scene. We further show that contextual cues serve as a powerful complementary signal for fake localization, improving SOTA detectors without any fine-tuning. Overall, \datasetName{} establishes a comprehensive testbed for advancing context-aware visual understanding, contextual reasoning, and image forensics. Future work could extend context reasoning to open-vocabulary settings beyond COCO's predefined categories, enabling more generalizable contextual understanding.

\section{Acknowledgments}
This work was funded in part by the National Science Foundation (SES-2521631) and the University of Georgia's Presidential Interdisciplinary Seed Grant. 

{
    \small
    \bibliographystyle{ieeenat_fullname}
    \bibliography{ref}
}


\end{document}